\documentclass[letterpaper, 10 pt, conference]{ieeeconf}  % Comment this line out if you need a4paper

\IEEEoverridecommandlockouts                              % This command is only needed if 
\usepackage{multirow} % multirow, multicolumn
\usepackage{graphicx} % includegraphic
\usepackage{svg} % includesvg
\usepackage{amsmath} % argmin
\usepackage{amsfonts} % mathbb
\def\mathbi#1{\textbf{\em #1}} % mathbi
\DeclareMathOperator*{\argmin}{argmin} % argmin
\usepackage{booktabs} % toprule, midrule, bottomrule
\usepackage{cite} % auto-style cite

\title{\LARGE \bf
Coarse-to-Fine Point Cloud Registration with SE(3)-Equivariant Representations
}

\author{Cheng-Wei Lin$^{1}$, Tung-I Chen$^{1}$, Hsin-Ying Lee$^{1}$,  Wen-Chin Chen$^{1}$, and Winston H. Hsu$^{1, 2}$% <-this % stops a space
\thanks{$^{1}$National Taiwan University, $^{2}$Mobile Drive Technology}
}

\begin{document}

\maketitle
\thispagestyle{empty}
\pagestyle{empty}

%%%%%%%%%%%%%%%%%%%%%%%%%%%%%% Main Paper %%%%%%%%%%%%%%%%%%%%%%%%%%%%%%%%

\begin{abstract}
Point cloud registration is a crucial problem in computer vision and robotics.
Existing methods either rely on matching local geometric features, which are sensitive to the pose differences, or leverage global shapes, which leads to inconsistency when facing distribution variances such as partial overlapping.
Combining the advantages of both types of methods, we adopt a coarse-to-fine pipeline that concurrently handles both issues. We first reduce the pose differences between input point clouds by aligning global features; then we match the local features to further refine the inaccurate alignments resulting from distribution variances.
As global feature alignment requires the features to preserve the poses of input point clouds and local feature matching expects the features to be invariant to these poses, we propose an SE(3)-equivariant feature extractor to simultaneously generate two types of features.
In this feature extractor, representations that preserve the poses are first encoded by our novel SE(3)-equivariant network and then converted into pose-invariant ones by a pose-detaching module.
Experiments demonstrate that our proposed method increases the recall rate by 20\% compared to state-of-the-art methods when facing both pose differences and distribution variances.
\end{abstract}

\section{Introduction}
% Overview
Point cloud registration, the task of finding rigid transformations that align two input point clouds, is fundamental for several applications in computer vision and robotics. Depending on the application domain, point cloud registration methods need to fulfill the demand for various properties \cite{bauer2021reagent}. For example, some applications such as SLAM \cite{kim2018slam, cattaneo2022lcdnet} require the registration method to be real-time and accurate. Other applications such as scene reconstruction \cite{takimoto2016reconstruction} expect the registration method to be robust to initial pose conditions.

\begin{figure}[!tp]
  \centering
  \includegraphics[width=0.48\textwidth]{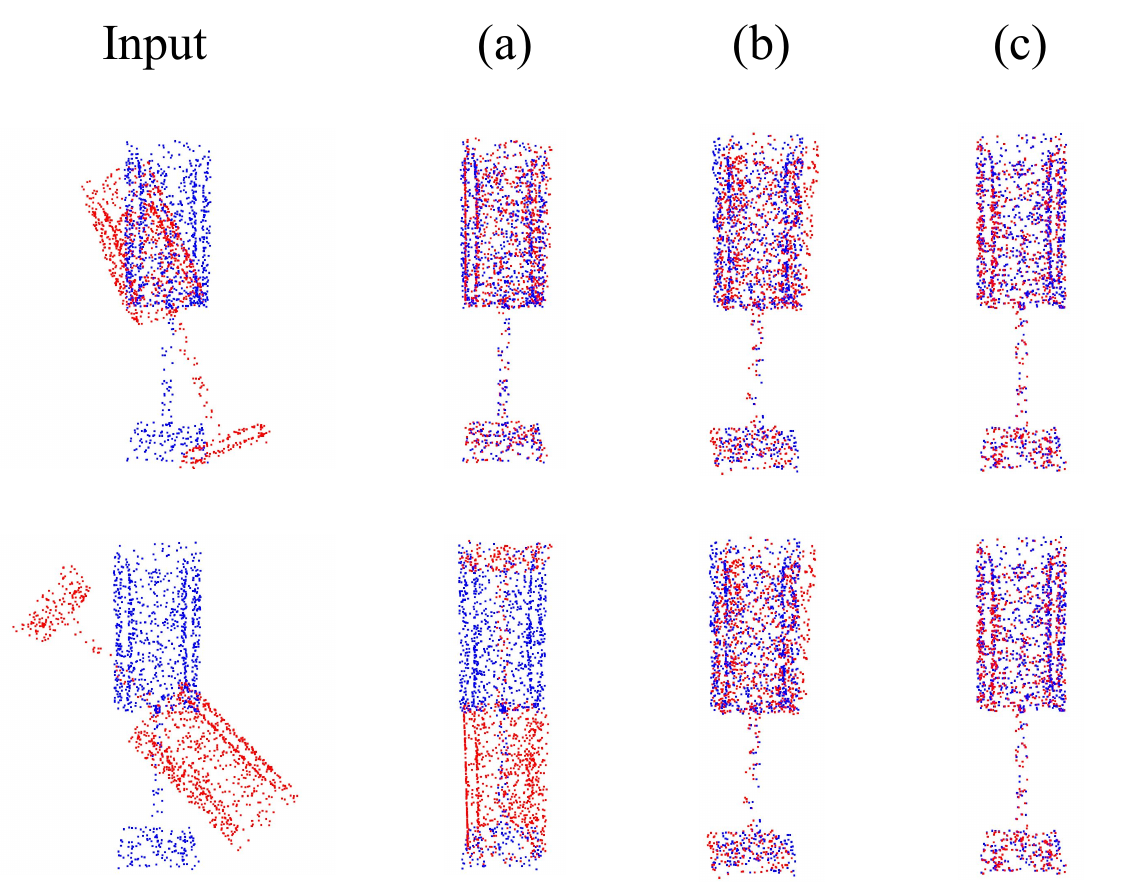}
  \caption{The registration results of methods from each category. We denote source point clouds as red and target point clouds as blue. (a) Local methods (e.g. IDAM) only produce accurate alignments when initial errors are small enough. (b) Global methods (e.g. DeepGMR) are invariant to initial errors but their performance are unreliable. In contrast, (c) our coarse-to-fine method accurately aligns point clouds regardless of initial errors.}
  \label{fig: overview}
\end{figure}

% Previous work fails to handle either large transformation or large distribution variances.
The diversified requirements lead to a variety of registration approaches.
Some prior approaches \cite{besl1992icp, wang2019dcp, li2020idam}, namely \textit{local} approaches, focused on time-efficiency and accuracy. However, due to the dependence on matching local geometric features, these approaches are sensitive to the magnitude changes in rigid transformations, and thus fail to handle large initial pose differences (Fig. \ref{fig: overview}-a). On the other hand, \textit{global} approaches \cite{yuan2020deepgmr, zhu2022correspondence} leverage global shape information to maintain robust against initial pose difference. `Nonetheless, these global approaches usually produce alignment results inferior to those of local approaches when facing \textit{distribution variances}, such as partial overlapping (Fig. \ref{fig: overview}-b). Distribution variances affect the overall shape that global methods rely on, while regional geometries remain unchanged.

\begin{figure*}[t]
    \vspace{4pt}
    \centering
    \includegraphics[width=1\textwidth]{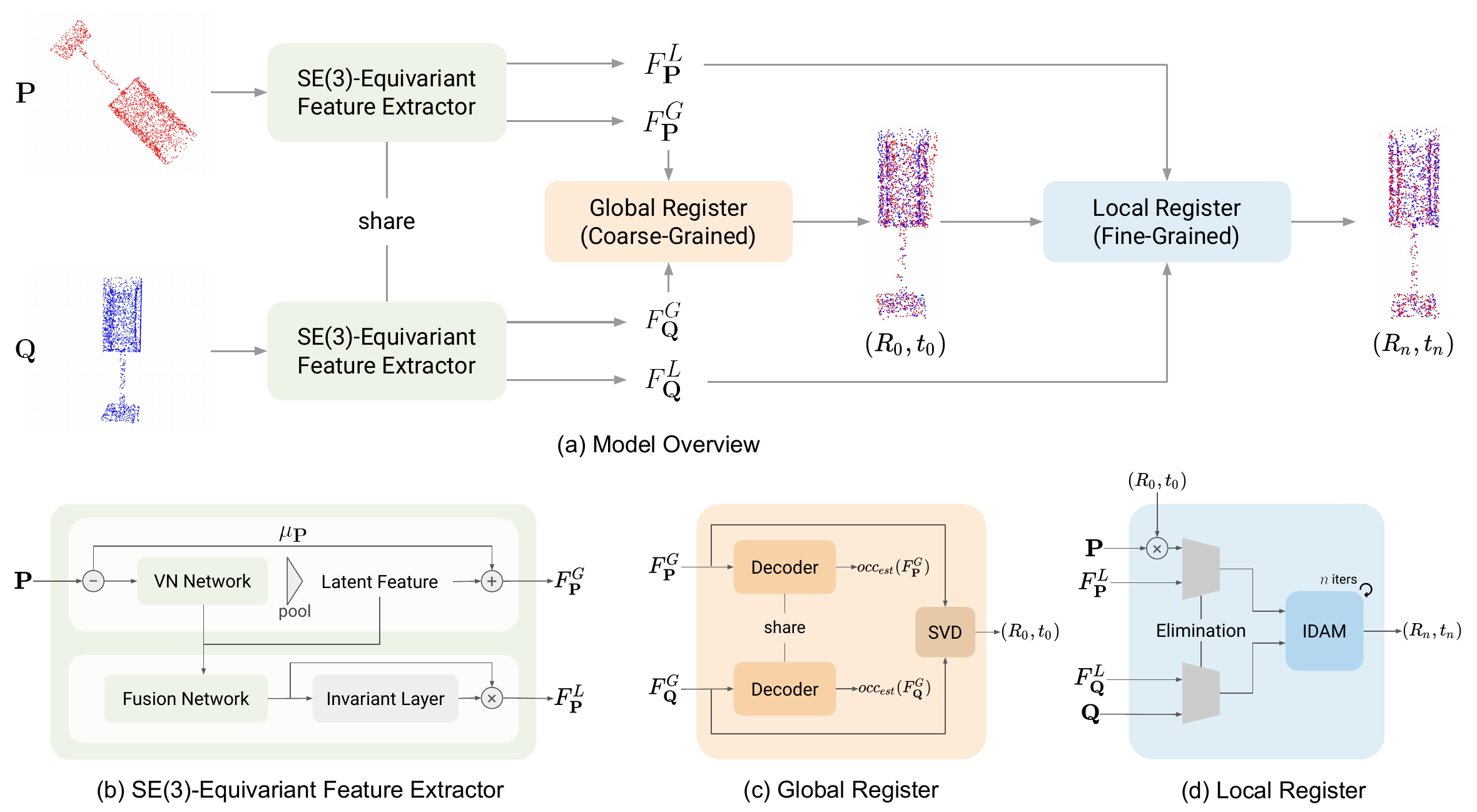}
    \caption{(a) An overview of the coarse-to-fine pipeline. Given point clouds $\mathbf{P}$ and $\mathbf{Q}$, (b) our SE(3)-equivariant feature extractor encodes SE(3)-equivariant global representations $F^G_\mathbf{P}$ and $ F^G_\mathbf{Q}$, as well as SE(3)-invariant local representations $F^L_\mathbf{P}$ and $ F^L_\mathbf{Q}$ (Sec. \ref{sec: feature extractor}). (c) As SE(3)-equivariant global representations preserve the poses of input point clouds, our global register produce rough alignments $(R_0, t_0)$ by aligning these global representations (Sec. \ref{sec: global register}). (d) Taken the roughly aligned point clouds and their SE(3)-invariant local representations, our local register further refines the alignment results by matching these local representations (Sec. \ref{sec: local register}). Consequently, our coarse-to-fine pipeline accurately aligns input point clouds by $(R_n, t_n)$. (Best seen in color.)}
    \label{fig: model architecture}
\end{figure*}

% We proposed a coarse-to-fine method to handle both cases.
To solve the registration problems with both large initial pose differences and distribution variances, we adopt a coarse-to-fine pipeline that takes advantage of both global and local approaches (Fig. \ref{fig: model architecture}-a).
A global register is applied as our coarse-grained register to reduce pose differences and roughly align input point clouds. Then, we utilize a local register to refine the inaccurate alignments caused by distribution variances.

% We introduced an SE(3)-equivariant feature extractor for the coarse-to-fine register.
On top of that, we employ a shared feature extractor to generate representations for both global and local registers. The feature alignment process in the global register requires representations to preserve the pose of input point clouds. On the contrary, the correspondence matching in the local register expects representations to get rid of the influences caused by the differences in the poses.

% How we build our feature extractor
In our feature extractor, an SE(3)-equivariant neural network and a pose-detaching module are proposed to produce pose-preserving features and convert them into pose-invariant ones, respectively (Sec. \ref{sec: preliminaries} shows more details about SE(3)-equivariance).
Unlike existing SE(3)-equivariant networks \cite{finzi2020generalizing, fuchs2020se3transformers}, our SE(3)-equivariant neural network avoids time-intensive approximations and kernels that constrain the expressiveness of the representation. We preserve the translations by maintaining the center of feature embedding and preserve the rotations using Vector Neuron \cite{deng2021vector}, a rotation-equivariant framework.
Furthermore, our pose-detaching module normalizes the input point clouds to remove the translations and utilizes the orthogonality of rotation matrices to eliminate the rotations.
Consequently, our novel SE(3)-equivariant feature extractor concurrently produces pose-preserving and pose-invariant representations, supporting both coarse- and fine-grained registers to effortlessly perform registrations.

% Experiments show that we outperform state-of-the-art under simulated real-world situations. Ablations also support our claim that we have overcome the distinct property challenge.
We evaluate our method on ModelNet40 \cite{wu2015shapenet}, which is composed of various object models. 
Following RPMNet \cite{yew2020rpm}, we pre-processed the dataset to simulate real-world situations, including sensor noises, independent scans, and partially overlapping point clouds. Furthermore, we evaluate the performance over multiple initial angle ranges to exhibit the influence of initial poses.
Experimental results demonstrate that our method outperforms state-of-the-art methods and reaches a reliable performance under simulated real-world scenarios. In addition, ablations (Sec. \ref{sec: ablations}) support that our feature extractor satisfies the feature requirements of both global and local registers yet remains time efficient.

% Summary
To sum up, the overall contributions of this work can be summarized as follows:
\begin{itemize}
    \item We apply a coarse-to-fine pipeline to resist impacts from both initial pose differences and distribution variances.
    \item We introduce a novel SE(3)-equivariant feature extractor, simultaneously obtaining representations for both global and local registers.
    \item Our method outperforms state-of-the-art methods on ModelNet40 under circumstances simulating the real world across different initial pose difference ranges.
\end{itemize}

\section{Related Work}
\subsection{Local Registration Methods}
Local approaches are often used under circumstances where transformations are known to be small in magnitude. Iterative Closest Point (ICP) \cite{besl1992icp} iteratively matches the closest points as correspondences and minimizes the distance between these correspondences, which often causes the result to converge at a local minimum. To resolve this problem, a variety of strategies have been proposed to deal with outliers \cite{chetverikov2005robust}, handle the noises \cite{fitzgibbon2003robust}, or devise better distance metrics \cite{low2004linear, segal2009generalized}. However, the limitation of matching points on Euclidean space leads to recent work performing matching on feature space. PPFNet \cite{deng2018ppfnet}, 3DSmoothNet \cite{gojcic20193dsmoothnet}, SpinNet \cite{ao2021spinnet}, and FCGF \cite{choy2019fcgf} follow this idea and solve the Procrustes problem based on the correspondences paired by their representations. Moreover, DCP \cite{wang2019dcp}, IDAM \cite{li2020idam}, RPMNet \cite{yew2020rpm}, DGR \cite{choy2020dgr}, ImLoveNet \cite{chen2022imlovenet}, and DetarNet \cite{chen2022detarnet} use the ground truth poses to supervise point matching and feature learning. Predator \cite{huang2021predator} and REGTR \cite{yew2022regtr} further leverage the ground truth overlap regions. Another branch of work such as D3Feat \cite{bai2020d3feat}, DeepVCP \cite{lu2019deepvcp}, PRNet \cite{wang2019prnet}, and GeoTransformer \cite{qin2022geometric} leverages key points to enhance the time efficiency. The remaining challenge is that perfect correspondences rarely exist in real-world situations, and thereby recent work \cite{yew2020rpm, drory2020best} utilizes soft matching \cite{gold1994softassign} to work under these conditions. Even so, these local methods still fail to handle large initial perturbations.

\subsection{Global Registration Methods}
Unlike local approaches, global approaches are designed to be invariant to the initial transformation error. Some methods such as GO-ICP \cite{yang2015goicp}, GOGMA \cite{campbell2016gogma}, and GOSMA \cite{campbell2019gosma} search the SE(3) space using branch-and-bound techniques. Other methods \cite{fischler1981random, horowitz2014convex, maron2016point} match the feature with robust optimization. However, these methods are unsuitable for real-time applications due to their large computation time. Fast Global Registration (FGR) \cite{zhou2016fgr} is presented to address this issue, achieving a similar speed to that of many local methods. To further improve the accuracy of the registration result, recent work handles the registration problem via learned global representations. DeepGMR \cite{yuan2020deepgmr} represents the global feature through GMM distributions and EquivReg \cite{zhu2022correspondence} takes rotation-equivariant implicit feature embeddings as its global representations. Nevertheless, these learning-based methods often struggle with distribution differences.

\subsection{Group Equivariant Neural Network}
Some research concentrates on proposing group equivariant neural networks as a means of resisting group transformations. For instance, Convolution Neural Network (CNN) \cite{lecun1998cnn} are translation-equivariant, resulting in its performance consistency among the same images with different 2D translations. To prevent the effect of rotation, recent studies \cite{cohen2016steerable, esteves2018so, fuchs2020se3transformers} construct the kernels by some steerable functions. However, these constrained kernels limit the flexibility of the network. Other studies \cite{kondor2018generalization, finzi2020generalizing} obtain the equivariance property by lifting the input space to higher-dimensional spaces where the group is contained. These studies are time-intensive and cost more computational resources due to the integration of the entire group. Vector Neuron \cite{deng2021vector} presents a brand new SO(3)-equivariant framework. The major advantage of this framework is the capability of incorporating the SO(3)-equivariance property into existing networks, such as PointNet \cite{qi2017pointnet} or DGCNN \cite{wang2019dgcnn}. We will later see how we design our SE(3)-equivariant feature extractor based on this simple idea, and use the extracted representation to cope with the registrations with notable initial transformations and distribution variances.

\section{Coarse-to-Fine Registration}
\label{sec: coarse-to-fine registration}
Illustrated in Fig. \ref{fig: model architecture}, our coarse-to-fine registration pipeline begins with extracting global and local feature representations. These global representations are fed into the global register to estimate a rough alignment between input point clouds. Focusing on roughly aligned point clouds, the local register refines the alignment results given the correspondences formed by matching these local representations.

\subsection{Preliminaries}
\label{sec: preliminaries}
A function $f \colon U \to V$ is equivariant to a set of transformations $G$, if for any $g \in G$, $f$ and $g$ commutes, i.e., $f(g\cdot u) = g \cdot f(u), \forall u \in U$. For instance, convolution layers are translation-equivariant because the outcome of applying a 2D translation to the input taken by convolution layers is identical to that of applying the 2D translation to the feature map resulting from the convolution layers. 
Therefore, we can say a network is SE(3)-equivariant if and only if the network $\textbf{nn}$ fulfills that for any pair of rotation $\mathbi{R} \in SO(3)$ and translation $\mathbi{t} \in \mathbb{R}^3$,
\begin{equation}
    \textbf{nn}(\mathbi{R} \mathbf{x} + \mathbi{t}) = \mathbi{R} \cdot \textbf{nn}(\mathbf{x}) + \mathbi{t},
\end{equation}
where $\mathbf{x}$ is the input 3D representations.  According to the above definition, our work is different from Neural Descriptor Fields \cite{simeonov2021neural}, which removes the rotations and translations rather than preserves them.

%%%%% Feature Extractor
\subsection{SE(3)-Equivariant Feature Extractor}
\label{sec: feature extractor}
To prepare for registration, we extract global features that represent the overall shape and local features that summarize the regional geometries. The feature alignment process in the global register expects the global features to preserve the SE(3) transformations of input point clouds. In contrast, the correspondence matching process in the local register requires the local features to avoid the impact of these transformations. To this end, our feature extractor comprises two modules, one managing to produce SE(3)-equivariant global representations $F^{G} \in \mathbb{R}^{C_0 \times 3}$ and the other aiming to generate SE(3)-invariant local representations $F^{L} \in \mathbb{R}^{N \times C_1}$.

    \subsubsection{SE(3)-Equivariant Global Features}
    \label{sec: global module}
    As the feature alignment process in the global register requires features to preserve the poses of input point clouds, we introduce a module to encode SE(3)-equivariant global representations.
    Depicted in the upper row of Fig. \ref{fig: model architecture}-b, this module is composed of three operations: maintaining the mean of input point clouds to preserve translations, passing through an SO(3)-equivariant network to preserve rotations, and average-pooling point-wise features to respect the fact that point clouds are not ordered.
    We maintain the mean by subtracting it at the beginning of the network and adding it back to the final global representation; hence preserve the translations. Next, we create a SO(3)-equivariant network by plugging Vector Neuron (VN) \cite{deng2021vector}, a rotation-equivariant framework, into a PointNet-based backbone, thus preserving the rotation and flexibly representing the attributes of each point. Lastly, the average-pooling approximates a permutation symmetry function, resulting in our global feature being permutation-invariant.
    % Due to the fact that average-pooling is also SO(3)-equivariant, the operation of feeding point cloud $\mathbf{X}$ into the module $f_G$ can be formulated as follow:
    % \begin{equation}
    %     f_G(\mathbf{X}) = f_{SO(3)}( \mathbf{X} \ominus \mu_\mathbf{X}) \oplus \mu_\mathbf{X},
    % \end{equation}
    % where $\mu_\mathbf{X} = \frac{1}{n}\sum_1^n{\mathbf{x}_i}$ and the SO(3)-equivariant network is denoted as $f_{so(3)}$. The operator $\ominus$ and $\oplus$ denotes element-wise subtraction and addition, respectively.
    % Consequently, the global representations encoded by this module are SE(3)-equivariant because of the VN-based neural network with the mean-recovery strategy. The proof is presented as follows:
    % \begin{equation}
    % \begin{aligned}
    % & f_{G}(\mathbi{R}\mathbf{X} \oplus \mathbi{t}) \\
    % & \quad = f_{SO(3)}(\mathbi{R}\mathbf{X} \oplus \mathbi{t} \ominus (\mathbi{R}\mu_\mathbf{X}+\mathbi{t})) \oplus (\mathbi{R}\mu_\mathbf{X} + \mathbi{t}) \\
    % & \quad = f_{SO(3)}(\mathbi{R}\cdot(\mathbf{X}\ominus\mu_\mathbf{X})) \oplus (\mathbi{R}\mu_\mathbf{X} + \mathbi{t}) \\
    % & \quad = \mathbi{R}\cdot f_{SO(3)}(\mathbf{X}\ominus\mu_\mathbf{X}) \oplus (\mathbi{R}\mu_\mathbf{X} + \mathbi{t}) \\
    % & \quad = \mathbi{R}\cdot(f_{SO(3)}(\mathbf{X}\ominus\mu_\mathbf{X}) \oplus \mu_\mathbf{X}) \oplus \mathbi{t} \\
    % & \quad = \mathbi{R}\cdot f_{G}(\mathbf{X}) \oplus \mathbi{t}.
    % \end{aligned}
    % \end{equation}
    With the above three operations, we encode feature representations that are SE(3)-equivariant and permutation-invariant.
    Denote the source point cloud as $\mathbf{P}$ and the target point cloud as $\mathbf{Q} = \mathbi{R}M\mathbf{P} + \mathbi{t}$, where $M$ is a permutation matrix. The relationship between their global representations $F^{G}_{\mathbf{P}}$ and $F^{G}_{\mathbf{Q}}$ is shown as follows:
    \begin{equation}
        F^{G}_{\mathbf{Q}} = \mathbi{R} \cdot F^{G}_{\mathbf{P}} + \mathbi{t}.
    \end{equation}

    \subsubsection{SE(3)-Invariant Local Features}
    \label{sec: local module}
    Additionally, we introduce another module to generate SE(3)-invariant local features for the local register to match correspondence points according to these features.
    Displayed in the lower row of Fig. \ref{fig: model architecture}-b, we follow PointNet \cite{qi2017pointnet} and fuse the global features into the local features by a VN-based fusion network.
    Then the fused feature embeddings $FR$ ($R$ denotes the preserved rotation) are made invariant to rotations by an invariant layer, which, based on VN, transforms $FR$ into vectors $VR$. With the operation of multiplying $FR$ by the transpose of $VR$, we obtain rotation-invariant local representations: 
    \begin{equation}
        (FR)(VR)^T = FRR^TV^T = FV^T.
    \end{equation}
    Moreover, these representations are also translation-invariant due to the mean subtraction mentioned in Sec. \ref{sec: global module}.
    As a result, these representations are invariant to any transformation from SE(3) group. That is, for point clouds $\mathbf{P}$ and $\mathbf{S} = \mathbi{R}\mathbf{P} + \mathbi{t}$, their local representations $F^{L}_{\mathbf{P}}$ and $F^{L}_{\mathbf{S}}$ are identical to each other.
    Extracting both SE(3)-equivariant global features and SE(3)-invariant local features, we fulfill the requirements of the following global and local registers.

%%%%% Coarse-Grained Register
\subsection{Global Register (Coarse-Grained)}
\label{sec: global register}
Given the SE(3)-equivariant global representations (Sec. \ref{sec: global module}), our coarse-grained register aims to reduce the pose differences in registration problems. These global representations have the same poses as the input point clouds; therefore we can simply solve the registration task by aligning their global representations when the point clouds only differ in their pose (Sec. \ref{sec: feature alignment}). To further deal with the noisy point clouds from different scans, we follow \cite{zhu2022correspondence} and employ the implicit representation loss (Sec. \ref{sec: occupancy loss}) as well as the registration loss (Sec. \ref{sec: registration loss}) to our training progress.

    \subsubsection{Feature Alignment}
    \label{sec: feature alignment}
    Our coarse-grained register operates registration on the global feature space since the global representations preserve the poses of input point clouds. Furthermore, these global representations are automatically paired due to the permutation-invariance property; thus the data association issue is removed, i.e., finding the correspondence point pairs between two point clouds is not necessary anymore. As a consequence, the registration problem of point clouds $\mathbf{P}$ and $\mathbf{Q} = \mathbi{R}M\mathbf{P} + \mathbi{t}$ can be solved by finding the rotation and translation between their global representations $F^{G}_\mathbf{P} = \{f^{G}_{\mathbf{P}, 1}, ..., f^{G}_{\mathbf{P}, c_0}\}$ and $F^{G}_\mathbf{Q} = \{f^{G}_{\mathbf{Q}, 1}, ..., f^{G}_{\mathbf{Q}, c_0}\}$, that is, optimizing the following objective function:
    \begin{equation}
        \argmin_{R, t}{ {\frac{1}{C_0}} \sum^{C_0}_{i=1} \Vert \mathbi{R}f^{G}_{\mathbf{P}, i} + \mathbi{t} - f^{G}_{\mathbf{Q}, i} \Vert ^2}.
    \end{equation}
    As validated in ICP \cite{besl1992icp}, this least square optimization can be solved in closed form by using single value decomposition.
      
    \subsubsection{Implicit Representation Loss}
    \label{sec: occupancy loss}
    The above feature aligning operation allows our coarse-grained register to align point clouds that differ in only permutations and rigid transformations, which is far from useful. In real-world applications, point clouds are captured from different scans, which do not result in identical sets of points. Inspired by EquivReg \cite{zhu2022correspondence}, we construct an encoder-decoder network, in which the encoder is the aforementioned SE(3)-equivariant feature extractor. The decoder estimates the occupancy value $occ_{est}(p;F^{G}_\mathbf{P}) \in [0, 1]$ of a queried position $p \in \mathbb{R}^3$ according to the encoded shape representation $F^{G}_\mathbf{P} \in \mathbb{R}^{C_0 \times 3}$ of point cloud $\mathbf{P}$. In practice, we query $n$ sampled positions for the implicit shape reconstruction and defined the loss of the occupancy value prediction as:
    \begin{equation}
        L_{occ} = \sum^n_{i=1} {\mathbf{cross\_entropy}(occ_{est}(p_i;F^{G}_\mathbf{P}), occ_{gt}(p_i))},
    \end{equation}
    where $occ_{gt}(p) \in \{0, 1\}$ is the ground truth occupancy value of the position $p$. With this network design, the global representations of point clouds captured from the same geometry are implicitly motivated to be similar.
    
    \subsubsection{Registration Loss}
    \label{sec: registration loss}
    Since these global representations are similar but not exactly the same, the slight difference among these point clouds may lead to considerable errors in registration outcomes. Hence, we include the registration step in our training progress to further encourage precise registration. With the estimated rigid transformation ($\mathbi{R}_{est}$, $\mathbi{t}_{est}$), we follow DCP\cite{wang2019dcp} and use the following registration loss to measure our coarse-grained register's agreement to the ground-truth rigid transformations ($\mathbi{R}_{gt}$, $\mathbi{t}_{gt}$):
    \begin{equation}
        L_{reg} = \Vert \mathbi{R}^T_{gt}\mathbi{R}_{est} - \mathbi{I}_3 \Vert^2_F +  \Vert \mathbi{t}_{gt} - \mathbi{t}_{est} \Vert^2_2,
    \end{equation}
    where $\Vert \cdot \Vert_F$ denotes the Frobenius norm and $\Vert \cdot \Vert_2$ denotes the Euclidean norm. Consequently, our coarse-grained register produces rough registration results $(\mathbi{R}_0, \mathbi{t}_0)$, which will be later refined by our fine-grained register.

%%%%% Fine-Grained Register
\subsection{Local Register (Fine-Grained)}
\label{sec: local register}
Since the original large-error registration tasks have been degraded into small-error ones with the global register (Sec. \ref{sec: global register}), we here present a local register that focuses on refining the small residual perturbation. Considering speed and simplicity, we apply \cite{li2020idam}, a learning-based local method, as our fine-grained register. As illustrated in \ref{fig: model architecture}-d, the network majorly consists of a hard elimination (Sec. \ref{sec: elimination}) and an Iterative Distance-Aware Similarity Matrix Convolution Network (IDAM) (Sec. \ref{sec: IDAM}).

    \subsubsection{Hard Elimination}
    \label{sec: elimination}
    To improve speed and filter out ambiguous regions, we employ an elimination network in the fine-grained register. According to the local representations, this elimination network gives every point a score and preserves the points along with their local representations of which the scores are in the top $1/6$. As claimed in \cite{li2020idam}, this process leaves the prominent points, such as corner points, and thereby simplifies the matching procedure later in IDAM.

    \subsubsection{IDAM}
    \label{sec: IDAM}
    Taking the eliminated points and representations, IDAM particularly learns two items, a similarity matrix and a weight vector. The similarity matrix presents the closeness of each source point and each target point, which can be used to find the corresponding points. The confidence in these correspondent points are approximated by the aforementioned weight vector. With the correspondent points and the confidence, a refined alignment result can be calculated by orthogonal Procrustes algorithms such as weighted SVD. Repeating this process $n$ times, we eventually get the accurate registration results $(\mathbi{R}_n, \mathbi{t}_n)$.

\section{Experiments}
\begin{table}[t]
    \vspace{4pt}
    \caption{Recall rates on clean data (Fig. \ref{fig: experiment settings}-a)}
    \label{tab: clean}
    \centering
    \begin{tabular}{lcccc}
    \toprule
        \multirow{2}{*}{Methods}& \multicolumn{4}{c}{Recall rates across different angle ranges} \\
        \cmidrule{2-5}
        & [0, 45] & [0, 90] & [0, 135] & [0, 180] \\
    \midrule
        IDAM \cite{li2020idam} & 99.3 & 98.2 & 96.9 & 97.0 \\
        RPMNet \cite{yew2020rpm} & 99.7 & 91.6 & 67.8 & 49.3 \\
        Predator \cite{huang2021predator} & 99.8 & 91.2 & 53.9 & 33.6 \\
    \midrule
        DeepGMR \cite{yuan2020deepgmr} & \textbf{100.0} & \textbf{100.0} & \textbf{100.0} & \textbf{100.0} \\
        EquivReg \cite{zhu2022correspondence} & \textbf{100.0} & \textbf{100.0} & \textbf{100.0} & \textbf{100.0} \\
    \midrule
        Ours & \textbf{100.0} & \textbf{100.0} & \textbf{100.0} & \textbf{100.0} \\
    \bottomrule
    \end{tabular}
\end{table}
\subsection{Dataset: ModelNet40}
\label{sec: dataset}
Our network is trained on ModelNet40 \cite{wu2015shapenet}, which consists of 12311 CAD models from 40 categories of objects. We follow the official split, taking 9843 models as the training set and the other 2468 models as the testing set. Besides, we use the tool provided by Stutz and Geiger \cite{Stutz2018watertight} to make these models watertight. Given the watertight models, we determine the occupancy value of any 3D coordinate depending on whether it is inside or outside the model. As for the transformations between input point clouds, we follow the procedure in DCP \cite{wang2019dcp}. That is, taking the specified maximum angle $\theta$ and maximum distance $d$ as arguments, our initial rotation and translation are determined by three Euler angles sampled in the range $[0, \theta]$ and three coordinates sampled in the range $[-d, d]$, respectively.

\subsection{Training and Testing Details}
\label{sec: implementation details}
Our training process is composed of two stages. First, we trained our coarse-grained register together with our feature extractor. Then, we fixed the global feature module and trained the fine-grained register. For evaluation, we set the maximum distance for the initial translation to $0.5$ across all settings. To demonstrate the robustness to initial pose differences, we tested the algorithms over initial rotations with maximum angles of $45^\circ$, $90^\circ$, $135^\circ$, and $180^\circ$.

\subsection{Evaluation Metrics}
\label{sec: evaluation metrics}
We evaluate our registration result by computing the evaluation metric provided by RPMNet \cite{yew2020rpm}:
\begin{equation}
    \mathbf{Err}(\mathbi{R}) = \arccos(\frac{tr(\mathbi{R}_{gt}^T \mathbi{R}_{est})-1}{2}),
    \mathbf{Err}(\mathbi{t}) = \Vert \mathbi{t}_{gt} - \mathbi{t}_{est} \Vert_2,
\end{equation}
where $\mathbf{Err}(\mathbi{R})$ denotes the rotation errors and $\mathbf{Err}(\mathbi{t})$ denotes the translation errors. In the following sections, we present the results as the recall rates of the rotation errors being less than $5^\circ$ and the translation errors being less than $0.2$.

\subsection{Baselines}
\label{sec: baselines}
To give readers a better context of differences between local and global registration approaches, we compared our method with some representative approaches in each category. IDAM \cite{li2020idam}, RPMNet \cite{yew2020rpm} and Predator \cite{huang2021predator} are the baselines representing local methods, while DeepGMR \cite{yuan2020deepgmr} and EquivReg \cite{zhu2022correspondence} are the baselines representing global methods. It's worth mentioning that since normal estimations are usually inaccurate in the real world \cite{yuan2020deepgmr}, we remove the input point pair features, which are calculated according to the normal, from RPMNet \cite{yew2020rpm}. Besides, we did not translate the input of the EquivReg \cite{zhu2022correspondence} because of its limited support for only rotational registration tasks. Its recall rates are hence formulated exclusively using rotation errors.

\begin{table}[t]
    \vspace{4pt}
    \caption{Recall rates on noisy data (Fig. \ref{fig: experiment settings}-b)}
    \label{tab: noisy}
    \centering
    \begin{tabular}{lcccc}
    \toprule
        \multirow{2}{*}{Methods}& \multicolumn{4}{c}{Recall rates across different angle ranges} \\
        \cmidrule{2-5}
        & [0, 45] & [0, 90] & [0, 135] & [0, 180] \\
    \midrule
        IDAM \cite{li2020idam} & 91.6 & 45.1 & 21.8 & 15.6 \\
        RPMNet \cite{yew2020rpm} & 91.4 & 62.0 & 33.8 & 25.5 \\
        Predator \cite{huang2021predator} & 98.0 & 59.6 & 21.8 & 10.7 \\
    \midrule
        DeepGMR \cite{yuan2020deepgmr} & 83.1 & 84.7 & 82.9 & 83.9 \\
        EquivReg \cite{zhu2022correspondence} & 89.0 & 89.0 & 88.9 & 89.3 \\
    \midrule
        Ours & \textbf{98.8} & \textbf{98.9} & \textbf{98.8} & \textbf{98.9} \\
    \bottomrule
    \end{tabular}
\end{table}
\begin{figure}[t]
    \centering
    \includegraphics[width=0.48\textwidth]{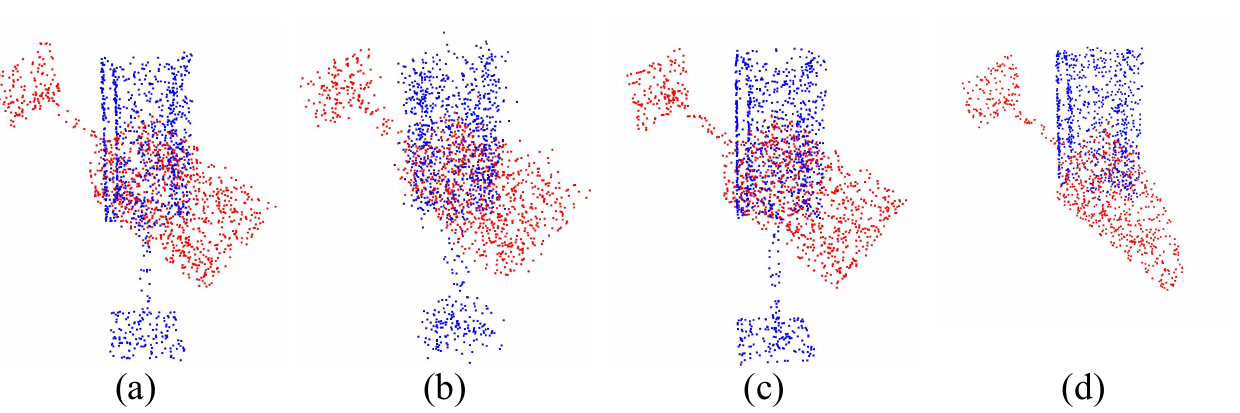}
    \caption{(a) Clean data: point clouds includes only permutation and pose difference. (b) Noisy data: point clouds are jittered by Gaussian noises, simulating sensor noises. (c) Independently sampled data: point clouds are separately sampled from the object surface, simulating independent scans. (d) Partially overlapping data: point clouds are cropped by farthest point sampling, simulating scans from different views.}
    \label{fig: experiment settings}
\end{figure}
\subsection{Performance on Multiple Point Cloud Settings}
As illustrated in Fig. \ref{fig: experiment settings}, we conducted our experiments with point clouds under multiple settings to simulate different real world scenarios, including clean, noisy, independently sampled, and partially overlapping point clouds.

    \subsubsection{Clean Data}
    \label{sec: clean data}
    We first test the registration of a point cloud and its permuted and transformed copy. The point clouds comprise 1024 points randomly sampled from the outer surface of each model. Under this setting, every point in the source point cloud has exactly one corresponding point in the target point cloud. Registration results are presented in Table \ref{tab: clean}. We can see that global methods and our method achieve recall rates of $100\%$ across different initial angle ranges. As for the local methods, IDAM performs appropriately while RPMNet barely handles small angle range cases.
    
    \subsubsection{Noisy Data}
    \label{sec: gaussian noise}
    To simulate the noises caused by the sensor, we further jittered the points in both point clouds with noises sampled from $N(0, 0.01)$ and clipped to $[-0.05, 0.05]$. As shown in Table \ref{tab: noisy}, local methods share a similar trend of recall rate decrement along with the increment of initial angle ranges. On the other hand, global methods reach high performance across different angle ranges. Therefore, our method can produce reliable coarse results and lead to even better results after the refinement.
    
    \begin{table}[t]
    \vspace{4pt}
    \caption{Recall rates on independently sampled data (Fig. \ref{fig: experiment settings}-c)}
    \label{tab: independently sampled}
    \centering
    \begin{tabular}{lcccc}
    \toprule
        \multirow{2}{*}{Methods}& \multicolumn{4}{c}{Recall rates across different angle ranges} \\
        \cmidrule{2-5}
        & [0, 45] & [0, 90] & [0, 135] & [0, 180] \\
    \midrule
        IDAM \cite{li2020idam} & 75.2 & 20.2 & 6.9 & 2.9 \\
        RPMNet \cite{yew2020rpm} &  90.5 & 58.4 & 22.1 & 11.2 \\
        Predator \cite{huang2021predator} & \textbf{99.5} & 85.9 & 46.4 & 25.2 \\
    \midrule
        DeepGMR \cite{yuan2020deepgmr} & 20.9 & 19.6 & 19.0 & 19.3 \\
        EquivReg \cite{zhu2022correspondence} & 43.8 & 44.0 & 44.4 & 42.6 \\
    \midrule
        Ours & 90.6 & \textbf{89.5} & \textbf{89.9} & \textbf{90.2} \\
    \bottomrule
    \end{tabular}
\end{table}
    \subsubsection{Independently Sampled Data}
    \label{sec: independently sampled}
    Instead of repeating points, real world scans generally contain distinct point sets despite being captured from the same view. Simulating these real world scenarios, we \textit{independently} sampled the 1024 points in source and target point clouds. The results in Table \ref{tab: independently sampled} reveal that global methods cannot handle these perturbations, and result in a significant performance drop. In contrast, local methods reach performances similar to those under conditions with Gaussian noises. On top of that, our method achieves consistent results regardless of the initial conditions, with only slightly poorer outcomes than those in Gaussian noise cases. Particularly, Predator reaches a better performance than ours on small initial angle ranges because the backbone of our fine-grained register is IDAM, whose results are inferior to those of Predator. We expect that changing the backbone of the fine-grained register will further enhance our performance.
    
    \begin{table}[t]
    \caption{Recall rates on partially overlapping data (Fig. \ref{fig: experiment settings}-d)}
    \label{tab: partially overlapping}
    \centering
    \begin{tabular}{lcccc}
    \toprule
        \multirow{2}{*}{Methods}& \multicolumn{4}{c}{Recall rates across different angle ranges} \\
        \cmidrule{2-5}
        & [0, 45] & [0, 90] & [0, 135] & [0, 180] \\
    \midrule
        IDAM \cite{li2020idam} & \textbf{98.8} & 69.4 & 37.6 & 27.1 \\
        RPMNet \cite{yew2020rpm} & 90.2 & 54.4 & 23.9 & 13.5 \\
        Predator \cite{huang2021predator} & 96.4 & 53.2 & 16.7 & 8.8 \\
    \midrule
        DeepGMR \cite{yuan2020deepgmr} & 46.7 & 48.2 & 48.2 & 46.6 \\
        EquivReg \cite{zhu2022correspondence} & 51.9 & 51.5 & 49.2 & 48.4 \\
    \midrule
        Ours & 82.1 & \textbf{79.3} & \textbf{78.0} & \textbf{78.3} \\
    \bottomrule
    \end{tabular}
\end{table}

    \subsubsection{Partially Overlapping Data}
    \label{partially overlapping}
    In most real-world applications, point clouds are captured from a single view, leading to incomplete shapes information on objects. To simulate these conditions, we followed RPMNet \cite{yew2020rpm} and sub-sampled the point clouds using the farthest point sampling algorithm, retaining approximately $75\%$ of the points. Table \ref{tab: partially overlapping} demonstrates that our method works properly under all initialization despite the slight degradation due to the information loss of the overall shape. Baseline methods either overfitted on cases with constrained angle ranges or failed to achieve reliable outcomes. Notably, local methods search for the answers in the constrained space rather than in the entire SE(3) space, which leads to their high performance in this constrained space but poor performance in spaces other than that.

\begin{table}[t]
    \vspace{4pt}
    \caption{Ablation study on equivariance and invariance}
    \label{tab: ablations}
    \centering
    \begin{tabular}{ccccc}
    \toprule
        VN & Invariant Layer & Shared Encoder & Recall(\%) & Speed(fps) \\
    \midrule
        & & & 17.9 & 17.8\\
        & $\checkmark$ & & 21.4 & 17.8\\
        $\checkmark$ & & & 73.2 & 17.9\\
        $\checkmark$ & $\checkmark$ & & 89.5 & 17.8 \\
        $\checkmark$ & $\checkmark$ & $\checkmark$ & \textbf{90.2} & \textbf{36.4}\\
    \bottomrule
    \end{tabular}
\end{table}
\subsection{Ablations}
\label{sec: ablations}
To clarify the efficacy and efficiency of our feature extractor, we presented the ablation studies in Table \ref{tab: ablations}. These ablation studies are evaluated under the independently sampled setting using a single Titan RTX GPU. 

Vector Neuron(VN) is used to generate SE(3)-equivariant global features, and the invariant layer is used to produce SE(3)-invariant local features. Results show that either non-equivariant global features or non-invariant local features will significantly decrease the recall rate. As a consequence, we confirm that global methods and local methods indeed expect their input feature representations to be SE(3)-equivariant and SE(3)-invariant, respectively. 

Table \ref{tab: ablations} also indicates that using dual encoders to extract these features will be time-consuming, and thus cannot be used in real-time applications. Our shared SE(3)-equivariant encoder, on the other hand, produces over $30$ fps without harming the efficacy. Moreover, we can further increase our speed by processing the local feature module (Sec. \ref{sec: local module}) and the global register (Sec. \ref{sec: global register}) in parallel.

\section{Conclusion}
In this paper, we proposed a coarse-to-fine point cloud registration method leveraging the representations extracted by an SE(3)-equivariant encoder. Our coarse-to-fine pipeline integrates the advantages of global and local registration methods, resulting in accurate alignment despite considerable initial transformations and severe distribution variances. Additionally, our SE(3)-equivariant encoder extracts the features fulfilling the necessity of the coarse-to-fine pipeline yet maintains the time efficiency required for real-time applications.

We conduct performance evaluations on data simulating the point clouds in real-world scenarios. Experiment results show that our network handles various input imperfections while still having room for improvement. For instance, when facing extreme noises, our coarse-grained register may fail to produce reliable results, which leaves substantial pose differences for our fine-grained register to handle. This problem is left for future work to solve.

\section{Acknowledgement}
This work was supported in part by National Science and Technology Council, Taiwan, under Grant NSTC 111-2634-F-002-022, and Mobile Drive Technology Co., Ltd (MobileDrive). We are grateful to the National Center for High-performance Computing.

%%%%%%%%%%%%%%%%%%%%%%%%%%%%%%%%%%%%%%%%%%%%%%%%%%%%%%%%%%%%%%%%%%%%%%%%%%%%%%%%


%%%%%%%%%%%%%%%%%%%%%%%%%%%%%%%%%%%%%%%%%%%%%%%%%%%%%%%%%%%%%%%%%%%%%%%%%%%%%%%%

% \clearpage

\bibliographystyle{IEEEtran}
\bibliography{reference}

\end{document}